\documentclass{ecai}
\usepackage{graphicx}
\usepackage{latexsym}
\usepackage{amsfonts}
\usepackage{xcolor}
\usepackage{tikz}
\usetikzlibrary{arrows.meta}
\usepackage{multirow, makecell}
\newcommand{\commentout}[1]{}
\usepackage{boldline}
\usepackage{enumitem}
\usepackage{calc}
\usepackage{newfloat}
\usepackage{listings}
\usepackage{pifont}
\newcommand{\xmark}{\ding{55}}%
\usepackage{amssymb}
\usepackage{tabularx}
\usepackage{lipsum,lmodern}
\usepackage{booktabs}
\usepackage[most]{tcolorbox}
\usepackage{algorithm}
\usetikzlibrary{positioning}
\usetikzlibrary{fit}
\usepackage{subcaption}

\usepackage[noend]{algpseudocode}
\algrenewcommand{\algorithmiccomment}[1]{\# #1}
\def\halfcheckmark{\checkmark\kern-1.1ex\raisebox{0.9ex}{\rotatebox[origin=c]{125}{--}}}

\newcommand{\ptype}[1]{\small\textsf{#1}}

\begin{document}

\begin{frontmatter}

\title{Automated Action Model Acquisition from Narrative Texts}


\author[]{
 Ruiqi Li$^{\spadesuit}$,  \ \
 Leyang Cui$^{\heartsuit}$,  \ \
 Songtuan Lin$^{\spadesuit}$, \ \
 Patrik Haslum$^{\spadesuit}$ \\
 $^\spadesuit$ Australian National University\\
 $^\heartsuit$ Tencent AI lab \\
 \textit{\{ruiqi.li,\ songtuan.lin,\ patrik.haslum\}@anu.edu.au} \quad\textit{leyangcui@tencent.com}\\

}



\begin{abstract}

Action models, which take the form of precondition/effect axioms, facilitate causal and motivational connections between actions for AI agents. Action model acquisition has been identified as a bottleneck in the application of planning technology, especially within narrative planning. Acquiring action models from narrative texts in an automated way is essential, but challenging because of the inherent complexities of such texts. We present NaRuto, a system that extracts structured events from narrative text and subsequently generates planning-language-style action models based on predictions of commonsense event relations, as well as textual contradictions and similarities, in an unsupervised manner. Experimental results in classical narrative planning domains show that NaRuto can generate action models of significantly better quality than existing fully automated methods, and even on par with those of semi-automated methods.
\end{abstract}

\end{frontmatter}

\section{Introduction}

In Artificial Intelligence (AI), planning comprises generating action sequences, i.e., plans, from the initial state of the problem to the goals. To construct such action sequences, AI planners require action models specified in a declarative planning language, such as the Planning Domain Definition Language (PDDL) \cite{mcdermott:PDDL:1998}. Building action models by hand is laborious and requires domain expertise. For narrative planning applications in particular (e.g., \cite{porteous-etal:aamas13}), due to the large variety of activities that occur in unstructured narrative texts, constructing action models for such narrative domains is challenging.

Recently, researchers have attempted to extract action models from narrative texts such as short stories and movie synopses \cite{hayton:etal:KEPS:2017,feng:ieee:2020,hayton:etal:AAAI:2020}. However, the methods proposed so far either generate quite simple and highly specific action models, or rely on human input to complement or correct automatic extraction.
Fully automated approaches have been applied to instructional texts such as recipes, manuals and navigational instructions \cite{mei2016listen,lindsay:etal:ICAPS:2017,feng-zhou-kambhampati:IJCAI:2018,olmo-sreedharan-kambhampati:arxiv:2021} (or, in some cases, transcriptions of plans generated from a ground truth domain into text). Such texts, however, lack many of the complexities of narrative texts, which are typically more colloquial and use complex clauses to express, e.g., conditional events. Hence, narrative texts (such as movie plots) are more difficult to comprehend. Therefore, how to automatically extract action models from narrative texts is still an open question.

In this paper, we introduce NaRuto (an abbreviation of ``narrative'' and ``automated''), an innovative, fully automated two-stage system for generating planning action models from narrative text. In the first stage, the system extracts structured representations of event occurrences from the source text, while in the second it constructs action models from predictions of commonsense event/concept relations.
NaRuto's distinctiveness lies in its ability to handle many complexities of narrative text, such as argument and conditional events, in employing commonsense knowledge inference, and maintaining full automation throughout the process.
We compare the action models generated by our system with those produced by prior approaches, where applicable, using two classical narrative planning domains. Results demonstrate that NaRuto creates action models of better quality than previous comparable fully automated methods, and sometimes even better than methods that incorporate manual input in the model creation process.

\begin{figure}[!t]
\centering
\resizebox{.9\columnwidth}{!}{\begin{tikzpicture}
    \node[draw=white] (input) {\footnotesize\textbf{Input}: Bryan hits Jack in the face.};
    \node[below left=1.5em and .5pt of input.west, anchor=north west] (output) {
        \footnotesize\textbf{Output}:
        \begin{minipage}[t]{.75\columnwidth}
            \begin{verbatim}
(:action hit
 :parameters   (?x - subject ?o - object)
 :precondition (and (close-to ?x ?o)
                    (angry-at ?x ?o)
                    (in-a-fight ?x ?o))
 :effect (and (yell-at ?o ?x)
              (injured ?o)
              (not (close-to ?x ?o))))
            \end{verbatim}
        \end{minipage}
    };
    \node[fit=(input)(output), draw, rounded corners=3pt]{};
\end{tikzpicture}}
\caption{Input: A narrative sentence. Output: The corresponding generated action model. 
}
\label{fig-intro-example}
\end{figure}


\noindent%
\paragraph{Overview of Our Approach} In narrative texts, events tell us what happened, who or what was involved, and at where. Actions are portrayed as events, so we extract events as the basis for generating actions. Event extraction is based on semantic role labelling (SRL) \cite{gildea-jurafsky:CL2002}, which identifies verbs and their arguments, as spans of text, and labels the arguments with their semantic role (e.g., agent, patient, modal modifier, etc). We complement SRL with several processing steps to refine events and their semantic relations.
In particular, there are two types of events that must be distinguished in order to accurately interpret the text:

\textit{Events as arguments}. Many verbs can take, or require, a clausal complement, i.e., an argument of the event verb is itself an event.
For example, consider this variation of the sentence in Figure \ref{fig-intro-example}: ``Bryan tries to hit Jack''. Here, the event verb is ``try'', and its argument is the event ``[Bryan] hit Jack''.
Clearly, the preconditions and effects of this event may differ from those shown in the example.
Event arguments can be nested: If ``Daniel sees Bryan try to hit Jack'', then the (complex) event ``Bryan try ([Bryan] hit Jack)'' is itself the argument of ``see''.
Since the occurrence of argument events depend on the main event, we regard the main and argument events as a whole by merging their verbs (e.g., ``try'' and ``hit'' become ``try to hit'').

\textit{Events as conditions}. Narrative texts frequently mention events that happen only if or when some condition(s) happens. For example, in ``She will hate me if I tell the truth.'', the event with the verb ``tell'' after ``if'' is a condition. Unlike argument events, such conditional events are not generally dependent on the conditioned event.
Therefore, we distinguish such conditional events from their consequent events and generate actions from them separately.

To generate action models from events, we make predictions of the preconditions and effects. In Figure~\ref{fig-intro-example}, for instance, being ``close to'' Jack is a precondition for Bryan to hit him in the face. Such preconditions and effects reflect the meaning of the event verb.
Although they may potentially be inferred from the verb's use across different contexts, we take a different approach and build a GPT~\cite{radford2018improving}-pretrained and BART~\cite{lewis2019bart}-finetuned commonsense event relation predictor (named COMET-BM) based on commonsense knowledge graphs describing everyday events \cite{sap2019atomic}.
COMET-BM works as a generative model: given a sentence describing an event and the desired relation type, it generates descriptions of concepts having that relation to the event. These are converted into preconditions and effects. To ensure that generated preconditions (resp. effects) are distinct and consistent, we further detect textual similarity and contradiction between them and eliminate those that are contradictory or similar to others.




\section{Background and Related Work}

\subsection{Action Model Generation From Narratives}
\label{sect:related-generation}


Extracting action models from text has recently gained
interest in AI planning. 
\commentout{\cite{riedl2010narrative,sil:yates:RANLP:2011,branavan2012learning,ware2014plan,yordanova2016textual,manikonda:etal:KEPS:2017,lindsay:etal:ICAPS:2017,hayton:etal:KEPS:2017,feng-zhou-kambhampati:IJCAI:2018,miglani:KEPS:2020,hayton:etal:AAAI:2020,jin2022integrating}.}
Sil and Yates \cite{sil:yates:RANLP:2011} identified web texts containing words that represent the target verbs based on their textual correlations and applied supervised learning methods to identify pre- and post-conditions for actions. Branavan et al. \cite{branavan2012learning} built a reinforcement learning model and used surface linguistic cues to learn possible action preconditions. Manikonda et al. \cite{manikonda:etal:KEPS:2017} extracted plan traces from social media texts to construct incomplete action models. 

Methods of action extraction vary, from using the dependency parse
structure to neural language models and reinforcement learning.
However,  Branavan et al. \cite{branavan2009reinforcement}, Yordanova \cite{yordanova2016textual}, Lindsay et al. \cite{lindsay:etal:ICAPS:2017}, Feng, Zhou and Kambhampati \cite{feng-zhou-kambhampati:IJCAI:2018}, Miglani and Yorke-Smith \cite{miglani:KEPS:2020}, and Olmo et al. \cite{olmo-sreedharan-kambhampati:arxiv:2021} focus on instructional texts, such as recipes, game-play instructions and user guides. Thus, they avoid much of the complexity that we face when dealing with narrative texts.
Hayton et al. \cite{hayton:etal:KEPS:2017} and Huo et al. \cite{feng:ieee:2020} considered narrative texts, but their action model generation is only partially automated, involving or even entirely relying on manual additions and corrections in the process. Hayton et al. \cite{hayton:etal:AAAI:2020} proposed a system that makes such inference procedure automatic, but the action models that they generate serve mainly to replicate the input narrative sequence, and do not generalize.

\subsection{Precondition and Effect Inference From Commonsense Knowledge Graphs}
\label{sect:related-commonsense}

Most events, or actions, mentioned in narratives are familiar ones. Their preconditions and physical and emotional effects are largely commonsense knowledge. Thus, we can look to commonsense knowledge (graphs) that incorporate events to infer them.

ConceptNet \cite{speer:AAAI:2019}
is a knowledge graph of concepts, which may be verb, noun or adjective phrases, and their relations, which brings together 3.4 million entity-relation tuples information collected from many sources, including
crowdsourced (e.g., DBPedia \cite{lehmann:dbpedia:2015}) and curated (e.g., OpenCyc, \cite{lenat:guha:1989}).
While it includes many relations that are relevant to modelling
actions/events -- for example, \textit{Causes}, \textit{HasSubevent},
\textit{HasPrerequisite}, and \textit{MotivatedByGoal} -- and over
128,000 concepts classified as verbs, the instances that have the relevant relations amount to only about 1\% of the graph \cite{davis2015commonsense,sap2019atomic}. For example, the number of verbs that have both \textit{HasPrerequisite} and \textit{Causes} -- i.e., both preconditions and effects -- is only 577.

The ATOMIC~\cite{sap2019atomic} knowledge graph is made up of 880K tuples linking 24K events to statements using 9 relations, including dynamic aspects of events like causes and effects, if-then conditional statements, social commonsense knowledge and mental states.
For example, the relation \textit{xNeed} holds between an event/action and a prerequisite for the action's subject to perform it, as in, for instance,
``X gets X’s car repaired'' \textit{xNeed} ``to have money''.
Crowdsourcing was used for both data collection and verification of this dataset.
Hwang et al.~\cite{hwang2021comet} presented ATOMIC-2020, a similar but more robust dataset. It is a new, high-quality commonsense knowledge graph comprising 1.33M commonsense knowledge tuples throughout 23 relations, encompassing social, physical, and event-based aspects of everyday inferential knowledge.  

%

We utilize these commonsense knowledge graphs to infer the pre- and post-conditions of action models because they reflect human experience and reasoning. These elements ensure that our model can create reliable action models from narrative texts, in an automatic manner.

\begin{figure*}[!t]
\centering

\includegraphics[width=\textwidth]{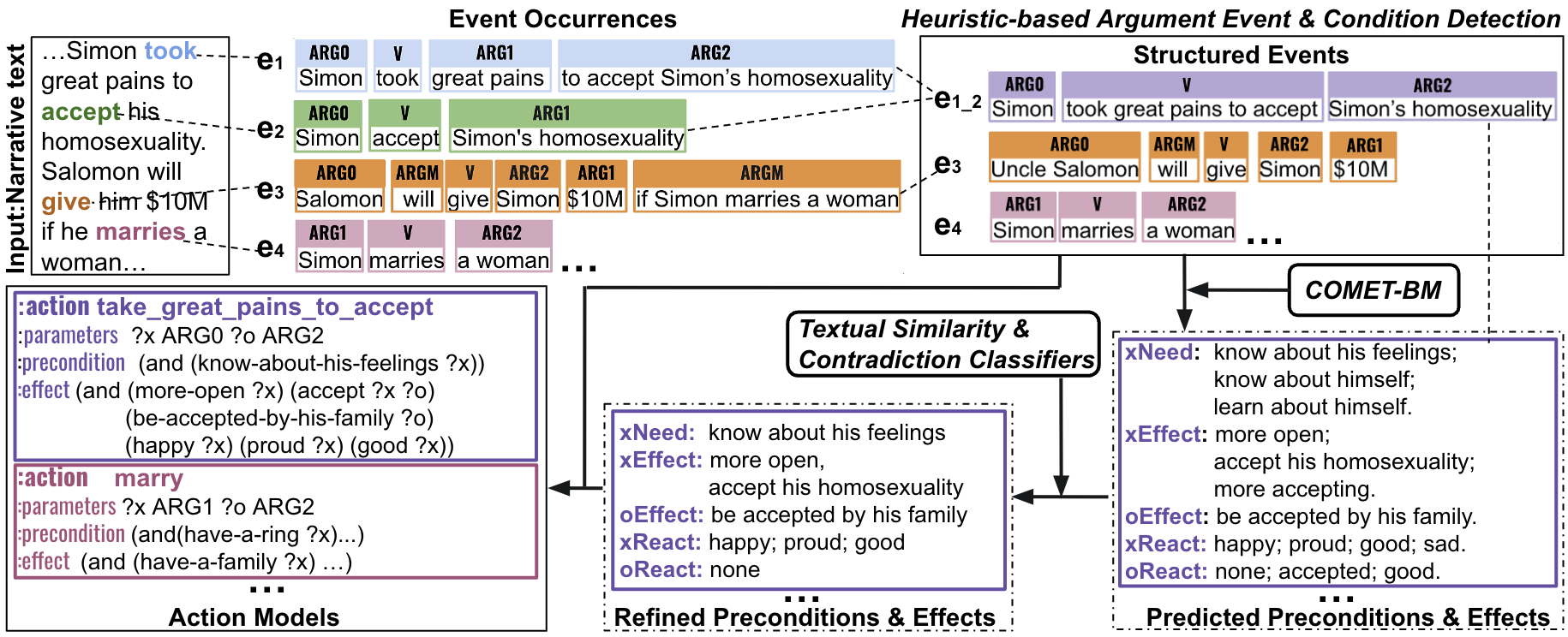}
\caption{Overview and example of our proposed approach for automatically generating action models from narrative text. The example input is part of a plot summary for the movie ``Man Is a Woman'', from
 Bamman et al. \protect\cite{bamman:movieplot:2013}.}
\label{fig-approach}
\end{figure*}

\section{Proposed Approach}

An overview of our proposed approach is shown in Figure~\ref{fig-approach}.
We extract event occurrences, consisting of verbs and their arguments,
using the AllenNLP \cite{allennlp:arxiv:2018} semantic role labelling (SRL) system, which is a BERT-based neural network~\cite{Shi:ArXiv:2019}.
We further use heuristic rules, based on dependency parse and POS tagging information, obtained using Stanford CoreNLP \cite{manning:ACL:2014}, for detecting phrasal verbs, argument events, and condition events, and update event structures accordingly (cf.\ section \ref{sect:extraction}).
We create action models from the events in two steps (cf.\ section \ref{sect:generation}): The first employs a commonsense event
relation predictor to generate candidate precondition and effect phrases;
in the second, these are filtered using textual similarity and contradiction,
so that both preconditions and effects are distinct and consistent.

\subsection{Structured Event Representation}
\label{sect:extraction}

\subsubsection{Event Verb and Arguments}
An event occurrence $e$ consists of a verb or phrasal verb $V(e)$ and a
set of labelled arguments $\mathbb{A}(e)$. Verbs are lemmatized.
We call any event whose verb lemma is ``be'' or ``have'' a
\emph{statement}, since these describe facts rather than events occurring, and we do not generate action models from them, except if an argument of the statement is an event.

The SRL system follows the PropBank annotation schema
\cite{bonial:journal:2012}, which divides argument labels into numbered
arguments (ARG0--ARG5), for arguments required for the valency of an
action (e.g., agent, patient, and so on), and modifiers of the verb,
such as purpose (PRP), locative (LOC), and so on.
Argument values are spans of text. Event Occurrences in Figure~\ref{fig-approach} illustrate extracted event arguments with their respective labels as colored chunks.

\subsubsection{Entity Resolution}

We apply co-reference resolution to the input narrative text, and substitute the first mention of any resolved entity for later mentions.
For example, in Figure~\ref{fig-approach}, ``his'' in the input text is substituted with the referenced entity ``Simon's'' in the events $e1$ and $e2$.
We use a document-level inference-based LSTM model~\cite{lee:coref:2018} from AllenNLP for the co-reference resolution task. 

\subsubsection{Phrasal Verb Detection}

Phrasal verbs are common in English, and identifying them is important
because their meaning is often
different from that of the verb part (e.g.,
``make up'' is different from ``make''; this is distinct from the fact
that ``make up'' also has several meanings).
The SRL system, however, extracts only single verbs.
We apply the following rule, adapted from \cite{komai:PACL:2015}, to detect
phrasal verbs:
If a word $P$ either (i) has a \textit{compound:prt} relation with the event verb $W$, or (ii) is adjacent to the event verb $W$ and has a \textit{case} or \textit{mark} relation with $W$ in the dependency parse tree, then $W P$ is a candidate phrasal verb; it is accepted if it appears in a list of known phrasal verbs\footnote{\url{https://en.wiktionary.org/wiki/Category:English_phrasal_verbs}}.


\subsubsection{Argument Event Detection}
We use the argument structure provided by the SRL system, together with a rule-based method that relies on the dependency parse information, to determine which events are arguments of other events.

Because arguments are spans of text, part or all of an extracted event
may lie within the argument of another event. If $V(e_j)$ is within an
argument of $e_i$, we say $e_j$ is \emph{contained} in $e_i$. This can
be nested.
Contained events are candidates for being arguments, but are not
necessarily so. For example, in Figure~\ref{fig-approach},
ARGM of $e_3$ contains $V(e_4)$ (``marries''),
but $e_4$ is not an argument of $e_3$.

We designed the following rules: If any of them is satisfied, a contained event $e_j$ is an argument of the containing event $e_i$:
(i) The dependency relation $V(e_i)$ to $V(e_j)$ is clausal complement
(\textit{ccomp} or \textit{xcomp}) or clausal subject (\textit{csubj}).
(ii) The dependency relation from $V(e_j)$ to $V(e_i)$ is copula (\textit{cop}) or auxiliary (\textit{aux:pass}).
(iii) All of $e_j$ is contained in an argument of $e_i$ that is labelled
with either ARGM-PRP (``purpose'') or ARGM-PNC (``purpose not cause'').

If an event $e_i$ has argument event(s), we take the span of all the event verbs and words within the range of the verbs as a verb phrase to be the updated $V(e_i)$. As shown in Figure~\ref{fig-approach}, $e_2$ is an argument event of $e_1$, so they are combined to $e_{1\_2}$, and $V(e_{1\_2})$ becomes ``take great pains to accept''.


\subsubsection{Condition Event Detection}

Conditional promises, threats, etc, are common in narrative text, as $e_4$ in Figure~\ref{fig-approach} shows. The condition event $e_4$ is not an argument of $e_3$, and should be removed from $e_3$. Hence, a different mechanism is required to identify conditions.

We use a method based on the signal words and phrases ``if'', ``whenever'',
``as long as'', ``on [the] condition that'', and ``provided that''.
For example, in Figure~\ref{fig-approach} the signal word ``if''
is in between the consequence $e_3$ and the condition $e_4$.
Our method is a modification of that introduced by
Puente et al. \cite{puente:ICFS:2010}.
Event $e_j$ is determined to be a condition of $e_i$ iff
(a) one of the sub-sequences $V(e_i) \quad S \quad V(e_j)$ or
$S \quad V(e_j) \quad V(e_i)$, where $S$ is one of the signal words/phrases, appear in the sentence, with no other (non-argument) event verb appearing in the sub-sequence;
and (b) one of the following holds:

\begin{list}{}{\leftmargin 0mm}
  \itemsep0em 
  \setlength{\labelwidth}{\widthof{\textbf{S5}:}}
  \settowidth{\itemindent}{\widthof{\textbf{S5}:}}
  \addtolength{\itemindent}{\labelsep}
\item[\textbf{S1}:] $V(e_i)$ is future simple tense, $V(e_j)$ is present simple;
\item[\textbf{S2}:] $V(e_i)$ is present simple tense, $V(e_j)$ is future simple;
\item[\textbf{S3}:] ``must'' or ``should'' or ``may'' or ``might'' is adjacent to $V(e_i)$, and $V(e_j)$ is present simple tense;
\item[\textbf{S4}:] ``would'' is adjacent to $V(e_i)$ and $V(e_i)$ is infinitive, $V(e_j)$ is past simple tense;
\item[\textbf{S5}:] ``could'' or ``might'' is adjacent to $V(e_i)$, and the tense of $V(e_j)$ is past simple;
\item[\textbf{S6}:] $V(e_i)$ is preceded by ``could'' and infinitive, and the tense of $V(e_j)$ is past continuous or past perfect;
\item[\textbf{S7}:] ``would have'', ``might have'' or ``could have'' is adjacent to $V(e_i)$ and $V(e_i)$ is a past participle, and $V(e_j)$ is past perfect tense;
\item[\textbf{S8}:] the tense of $V(e_i)$ is perfect conditional continuous, the tense of $V(e_j)$ is past perfect;
\item[\textbf{S9}:] the tense of $V(e_i)$ is perfect conditional, the tense of $V(e_j)$ is past perfect continuous;
\item[\textbf{S10}:] ``would be'' is adjacent to $V(e_i)$ and the form of $V(e_i)$ is gerund, $V(e_j)$ is past perfect;

\end{list}
The tenses of event verbs are determined by their POS tags.
The updated event structures after detecting argument and conditional events are shown in Figure~\ref{fig-approach} Structured Events.

\subsection{Action Model Creation}
\label{sect:generation}

An action model consists of four components: the action name, parameters, preconditions, and effects.
We generate an action from each structured event, using the event verb as the action name, or the combined verb phrase if the event has argument events, as in the example in Figure \ref{fig-approach}.
Each action has one or two parameters, \textit{x} and \textit{o}, representing the subject and (direct) object of the event, selected from the event argument based on the SRL argument labels (cf.\ section \ref{sect:generation-parameters}).
We trained a commonsense event/concept relation predictor (COMET-BM) to generate candidate preconditions and effects from the event text (cf.\ section \ref{sect:generation-prediction}). Candidate preconditions and effects are filtered to remove semantic duplicate and contradictory phrases, to ensure they are consistent. We also use textual entailment to infer negated effects and preconditions (cf.\ section \ref{sect:generation-selection}).

\subsubsection{Precondition and Effect Prediction}
\label{sect:generation-prediction}

Our event/concept relation predictor is trained on the three datasets
ATOMIC~\cite{sap2019atomic}, ATOMIC-2020~\cite{hwang2021comet} and ConceptNet~\cite{speer:AAAI:2019}, introduced in Section \ref{sect:related-commonsense}.

We adopt COMET \cite{bosselut2019comet}, which is a GPT
model \cite{radford2018improving} that is pre-trained on the ATOMIC and the ConceptNet knowledge graphs, and build a BART~\cite{lewis2019bart}-based variation (named COMET-BM) by finetuning it on a subset of the ATOMIC-2020 dataset.
Specifically, we select triples involving the relations
\textit{xNeed} (precondition for the subject, \textit{x}, to undertake
or complete the event), \textit{xEffect} (effect on the subject, \textit{x},
of the event), \textit{oEffect} (effect on the object, \textit{o}, of the
event), \textit{xReact} and \textit{oReact}
(reaction, i.e., emotional effect, on the subject, \textit{x}, and object,
\textit{o}, respectively).
Together, there are over 472K instances of these relations in the ATOMIC-2020 dataset.
In the COMET-BM training procedure, we construct the tuple <$I, T$> from the dataset. $I$ denotes the concatenation of $e$ and $r$, where $e$ is the text describing the event (verb and argument spans) and $r$ is the relation. $T$ is the textual description denoting the commonsense knowledge inferred from $I$. Following Bhagavatula et al.~\cite{bhagavatula2019abductive}'s work, we add two special tokens \texttt{<s>} and \texttt{</s>} to mark the beginning and the end for each $I$. The conditional probability of the $n$th token of $T$ is defined as:

\begin{equation*}
    P(T_n | T_{[0, n-1]}) = Softmax(W*D(H_{T_{[0, n-1]}}, E(I)) + b),
\end{equation*}
where $T_n$ and $T_{[0, n-1]}$ are the $n$th token and all preceding $(n-1)$ tokens in $T$; $E$ and $D$ are the encoder and decoder of the COMET-BM model (details refer to the BART model structure); $H_{T_{[0, n-1]}}$ is the decoded hidden states of all $n-1$ tokens; and $W$ and $b$ are learnable weight and bias parameters, respectively.
During training, the objective function for COMET-BM to minimize is the negative log-likelihood: 
\begin{equation*}
    \mathcal{L} = - \sum_{n=1}^{|T|} \log P(T_n | T_{[0, n-1]})
\end{equation*}


Phrases that have the \textit{xNeed} relation with the event become
candidate preconditions, and the others candidate effects.
Because action can have multiple preconditions and effects,
COMET-BM works as a text generation model: it takes the event text $e$ and relation $r$ as input and outputs the candidate phrase $I$ with an unfixed number of words.
For each $e$, $r$, we generate up to $K$ different outputs with the highest probabilities, from which we retain the ones with a normalized probability greater or equal to a threshold $\theta_r$.
Based on the probability distributions of each relation's predictions, we
set $K = 6$, $\theta_{\textit{xNeed}}$ to $0.7$,
$\theta_{\textit{xEffect}}$ and $\theta_{\textit{oEffect}}$ to $0.5$, and
$\theta_{\textit{xReact}}$ and $\theta_{\textit{oReact}}$ to $0.2$.
If any of the predicted phrases is ``none'', all lower-probability predictions are omitted. 
As an example, Figure~\ref{fig-approach} Predicted Preconditions \& Effects shows the top predictions for each relation, given $e_{1\_2}$ as input. 



\subsubsection{Precondition and Effect Selection}
\label{sect:generation-selection}

Because COMET-BM generates each candidate precondition and effect
separately, when taken together they are not necessarily semantically distinct or consistent.
In Figure~\ref{fig-approach}, for example, the two predicted preconditions
``know about his feelings'' and ``know about himself'' for event $e_{1\_2}$ are semantically similar, and including both is redundant.

Hence, we filter COMET-BM outputs by deleting lower-probability
preconditions that contradict or are similar to ones of higher probability,
and likewise for effects.
We pair the phrases output by COMET-BM for each of the (up to) six relations, and use two sentence-transformer \cite{reimers-2019-sentence-bert} models to predict if they are similar or contradictory. If yes, we eliminate the one with a lower probability.

The similarity predictor\footnote{\url{https://huggingface.co/sentence-transformers/all-MiniLM-L6-v2}} is based on a large pre-trained model for natural language understanding and finetuned on over 1B phrase pairs. It computes the cosine similarity of 384-dimensional phrase embeddings. We judge two phrases to be redundant if their similarity is $0.5$ or higher.
The second model\footnote{\url{https://huggingface.co/cross-encoder/nli-deberta-v3-base}} is finetuned on over 4M sentence pairs from the SNLI \cite{bowman-etal-2015-large} and MultiNLI \cite{williams:2018:mnli} datasets, and ranks whether the relation between two phrases $(a,b)$ is most likely to be entailment (i.e., $a$ implies $b$), neutral ($a$ and $b$ are logically independent) or contradiction.
Both models have a reported test accuracy of over 90\%.

We also use the textual contradiction classifier to generate negated action effects and preconditions. If a literal \ptype{($p$ ?x ?o)} is contradicted by a positive effect (resp. precondition), then \ptype{(not ($p$ ?x ?o))} is a candidate negative effect (resp. precondition).
We have tried two generating strategies: in full negation (named \textit{global}), we apply this test to all predicates defined in the domain; in restricted negation (named \textit{local}), we generate only effects that are negations of literals that appear in the action's precondition, and no negated preconditions.



\subsubsection{Parameter Selection}
\label{sect:generation-parameters}

The commonsense relation predictor expects each event to have a subject, \textit{x}, and optionally an object, \textit{o}. Hence, each generated action has one or two corresponding parameters. These are selected from the event arguments, based on the SRL labels, following Algorithm~\ref{alg:parameter}. Only arguments with numbered labels ($ARG0$--$ARG5$), which represent the event's agent, patient, and so on, are considered; event modifiers detected by SRL can not instantiate parameters.

\renewcommand{\algorithmicrequire}{\textbf{Input:}}

\begin{algorithm}
\caption{Extracting Parameters from Event Arguments.}\label{alg:parameter}
\begin{algorithmic}[1]
\Require $e$ \quad \Comment{event}
\Procedure{}{}
\State $x = \mathrm{null}$ \quad  \Comment{subject}
\State $o = \mathrm{null}$ \quad \Comment{object}
\State $ \mathbb{A}(e) = \{(a_1, lbl_1), \ldots, (a_k, lbl_k)\}$ \quad  \Comment{numbered arguments in $e$}
\State \Comment{Stage 1: Find subject}

\If{$(a, ARG0) \in \mathbb{A}(e)$}
\State $x = a$
\ElsIf{$(a, ARG1) \in  \mathbb{A}(e)$}
\State $x = a$
\EndIf
\State \Comment{Stage 2: Find object}
\If{$x \neq \mathrm{null}$}
\For{$i \in [s+1,5]$}
\If{$(a, ARGi) \in  \mathbb{A}(e)$}
\State $o = a$
\State $\textbf{break}$
\EndIf
\EndFor
\EndIf
\State \textbf{return} $x$ and $o$
\EndProcedure
\end{algorithmic}
\end{algorithm}

The action parameter \ptype{?x} becomes an argument of each predicates obtained from the \textit{xNeed}, \textit{xEffect} and \textit{xReact}
relations; likewise \ptype{?o} becomes an argument of predicates
obtained from the \textit{oEffect} and \textit{oReact} relations.
However, if the argument that instantiates the other parameter in the
event appears, literally, in the predicate, it is also removed
and replaced with the other parameter. For example, given the prediction
``X gets X’s car repaired'' \textit{xReact} ``X doesn't like X's car'',
the generated effect will be 
\ptype{(doesnt\_like ?x ?o)}.


\section{Evaluation}
To evaluate NaRuto, we want to determine how well the action models generated from a narrative text describe the intended meaning of the actions, and compare this with models obtained by previous systems that address the same task\footnote{Evaluations of the performance of some individual components of the system can be found in the supplementary material.}.
Let us start with why this is not easy.
First, action verbs extracted from text often have multiple meanings; even the same action may have different conditions or effects depending on the narrative context.
For example, in a fantasy story, a knight who slays a dragon will often become a hero -- unless the story is told from the point-of-view of a den of dragons.
Second, the same meaning can be expressed in different ways, using different predicates. For example,
the condition ``\ptype{?x} and \ptype{?y} are in the same place'' could be a binary predicate \ptype{(in-same-place ?x ?y)}, or the two facts ``\ptype{(at ?x ?place)}'' and ``\ptype{(at ?y ?place)}''.
Third, although quite a few approaches to action sequence extraction and model generation from text have been proposed, only a small number target narrative text, and have results available for comparison.

For input, we use two short stories that have appeared in work on narrative planning: Riedl and Young's \cite{riedl2010narrative} Aladdin story, and Ware's \cite{ware2014plan} Old American West story. Both are hand-written textual descriptions of plans generated by narrative planning systems. The advantages of using these texts are that there thus exists a ground truth, in the form of the planning domains used by the respective narrative planners, and that these two stories have also been used for evaluation in previous work on action model learning from narrative text \cite{hayton:etal:KEPS:2017,feng:ieee:2020}, so that some results are available for comparison.
A downside is that these texts are somewhat simpler than what we intend NaRuto to target. For example, neither story mentions any conditional event.

We compare the action domain models generated by NaRuto to those generated by Hayton et al's 2017 StoryFramer system \cite{hayton:etal:KEPS:2017}, and their 2020 system \cite{hayton:etal:AAAI:2020} (abbreviated ``H2020'').
Neither system is available for use, but the domains produced by StoryFramer for the two example stories were kindly provided by the authors, and we have attempted to replicate the results of H2020 applied to the same stories using the StoryFramer material and the description in their paper.
The details are described in section \ref{sect:baselines} below.
Huo et al. \cite{feng:ieee:2020} provide some results for Ware's Old American West story, but not a complete domain model; we compare the aspects of NaRuto that we can with their data.

We focus our evaluation on the set of actions modelled.
We do not try to align the predicates defined in each of the different domain models generated from the same text.
As noted, models can describe the same actions with different sets of predicates, or equivalent predicates that have different names. Furthermore, the narrative text usually focuses on what happens, i.e., the events, and only infrequently mentions what is, i.e., facts. Thus, we find higher agreement between different approaches in the set of actions.
We compare the sets of action names extracted from each story with those in the ground truth domain, using manual alignment of names. The same comparison was also done by Hayton et al. \cite{hayton:etal:KEPS:2017} and Huo et al. \cite{feng:ieee:2020}. The results are presented in section \ref{sect:identification} below.
Furthermore, we conducted a (blind) expert assessment of the appropriateness of each action's preconditions and effects, relative to the predicates defined in the domain.
The details of this evaluation process and its results are described in section \ref{sect:subjective} below.


\subsection{Implementation Settings}

NaRuto was run on a computer with 64 CPUs with 126GB of RAM and one RTX-3090 GPU with 24GB of RAM. We use stanford-corenlp-4.2.0 version toolkit for dependency parsing and POS tagging. We train our COMET-BM model on this single GPU, which takes around 4 hours. We set the optimization method as AdamW~\cite{loshchilov2019adamw} with an initial learning rate of $1\mathrm{e}{-5}$, a batch size of 32, and 3 epochs during the training (fine-tuning) process.

\subsection{Comparison Systems}\label{sect:baselines}

StoryFramer \cite{hayton:etal:KEPS:2017} is a partially automated domain modelling tool. Given a narrative source text, it automatically extracts candidate action verbs, candidate predicates based on properties, and candidate objects (nouns). The remaining modelling task is left to the user, who must assign types to objects and action parameters, identify duplicates, and, crucially, select which predicates to use as preconditions and effects for each action. The system adds some default predicates and preconditions, such as automatic inequality constraints for all action parameters of the same type. The user can also override or edit any system suggestion (e.g., add/remove candidate predicates, objects, etc).
Hayton et al. \cite{hayton:etal:KEPS:2017} applied StoryFramer to the Aladdin and Old American West stories. While we cannot replicate their process, both because the system is not available for use, and because we do not know what edits the user(s) in that study made, the resulting domain files were provided by the authors.

Their recent system \cite{hayton:etal:AAAI:2020} (``H2020'') is automatic. It is also not available for us to use, but since its event extraction mechanism is very similar to that of StoryFramer, apart from using a novel co-reference resolution method, we approximate its result by supposing it would extract the same action signatures (names and parameters) as StoryFramer, and constructing the corresponding action models following the method described in the paper.
The action models created by H2020 aim to replicate the input narrative sequence.

Huo et al. \cite{feng:ieee:2020} propose a partially automated system to learn a planning domain model, applied to natural disaster contingency plans. They use POS tagging to extract (action name, subject, object) triples, representing actions taking place. Their system also involves a human user to refine the action model. Because neither the system nor its outputs are available to us, we can only compare with the results included in their paper.

Ware's \cite{ware2014plan} Old American West text consists in fact of several story variants; since H2020 depends on the order of events in the source text, we apply it to one of them (plan G, the longest). The other systems use the concatenation of all story variants as inputs.

\subsection{Identification of Narrative Actions}
\label{sect:identification}

Table~\ref{tab-action-compare} lists the action names extracted by StoryFramer \cite{hayton:etal:KEPS:2017}, Huo et al.'s \cite{feng:ieee:2020} system and by NaRuto, and compares them with actions defined in the ground truth, i.e., the hand-written narrative planning domains from Ware's thesis~\cite{ware2014plan} (West story) and from Riedl and Young~\cite{riedl2010narrative} (Aladdin). Results for StoryFramer and Huo et al.'s \cite{feng:ieee:2020} system are from the respective papers. Note that Huo et al.\ did not try their system on the Aladdin story.


\begin{table}
\small
\begin{center}
\caption{Action names extracted from the two input stories by StoryFramer \protect\cite{hayton:etal:KEPS:2017}, Huo et al.'s system
\protect\cite{feng:ieee:2020}, and NaRuto.
The first column (GT) lists the corresponding action names in the ground truth, i.e., the hand-written narrative planning domain files by Ware~\protect\cite{ware2014plan} (Old American West story) and Riedl and Young ~\protect\cite{riedl2010narrative} (the Tale of Aladdin). \textcolor{red}{\xmark} means the action is not detected; \textcolor{red}{\halfcheckmark} means the action is partially extracted or incomplete; \textcolor{red}{$\ast$} means the action is not present in the ground truth planning domain but occurs in the story text.
}


\resizebox{\columnwidth}{!}{\begin{tabular}{llll} 
\textbf{GT~\cite{ware2014plan}/\cite{riedl2010narrative}} & \textbf{StoryFramer~\cite{hayton:etal:KEPS:2017}} & \textbf{Huo et al.~\cite{feng:ieee:2020}} & \textbf{NaRuto}\\
\toprule
\multicolumn{4}{l}{Domain 1: An Old American West Story}                                                                                                       \\
\midrule
die                             & \checkmark died                                 & \checkmark died                          & \checkmark die                   \\
heal                            & \checkmark healed                               & \checkmark healed                        & \checkmark heal                  \\
shoot                           & \checkmark shot                                 & \checkmark shot                          & \checkmark shoot                 \\
steal                           & \checkmark stole                                & \checkmark stole                         & \checkmark steal                 \\
snakebite                       & \checkmark bitten                               & \textcolor{red}{\halfcheckmark} got      & \checkmark get bitten            \\
                                & \textcolor{red}{\halfcheckmark} intended        & \checkmark intended to heal              & \checkmark intend to heal        \\
                                &                                                 & \checkmark intended to shoot             & \checkmark intend to shoot       \\ 
                                &                                                 & \textcolor{red}{$\ast$} using            & \textcolor{red}{$\ast$} use      \\ 
                                &                                                 & \textcolor{red}{$\ast$} angered          & \textcolor{red}{$\ast$} anger    \\ 
\midrule
\multicolumn{4}{l}{Domain 2: The Tale of Aladdin}                                                                                                               \\
\midrule
travel                          & \checkmark travels                              &    \multirow{13}{*}{}                   & \checkmark travel\\
slay                            & \checkmark slays                                &                                                    & \checkmark slay\\
pillage                         & \checkmark takes                                &                                                    & \checkmark take\\
give                            & \checkmark gives                                &                                                    & \checkmark hand\\
summon                          & \textcolor{red}{\xmark}                         &                                                    & \checkmark summon\\
love-spell                      & \checkmark casts                                &                                                    & \checkmark cast\\
fall-in-love                    & \textcolor{red}{\xmark}                         &                                                    & \checkmark fall-in \\
marry                           & \checkmark wed, married                         &                                                    & \checkmark wed\\

                                & \textcolor{red}{$\ast$} confined                &                                                    & \textcolor{red}{$\ast$} be-confined\\
                                & \textcolor{red}{$\ast$} rubs                    &                                                    & \textcolor{red}{$\ast$} rub\\
                                & \textcolor{red}{$\ast$} sees                    &                                                    & \textcolor{red}{$\ast$} see\\ 
                                &                                                &                                                     & \textcolor{red}{$\ast$} make\\
                                &                                                 &                                                    & \textcolor{red}{$\ast$} be-not-confined\\
\bottomrule
\end{tabular}}


\label{tab-action-compare}
\end{center}
\end{table}

There is not a perfect match between action names defined in the ground truth domains and those extracted from the narrative texts because the texts use different words to describe them; for example, the action \ptype{give} in the Aladdin story is described as ``Aladdin hands the magic lamp to King Jafar'', so the automatically extracted action name is \ptype{hand}.
For Ware's \cite{ware2014plan} Old American West story, the events ``use'' and ``anger'' are not actions in the ground truth planning domain, but are implicitly represented in the effects of other actions, e.g., the action \ptype{heal} requires the character who performs it to have medicine, which is used up as part of the action's effects. These events occur in descriptions of those actions, in the story sentences ``Carl the shopkeeper healed Timmy using his medicine'' and ``Hank stole antivenom from the shop, which angered sheriff William''.
Our detection of argument events is seen in, for example, the actions ``get bitten'' or ``intend to shoot'', where StoryFramer only extracts ``intended'' from the sentence ``Sheriff William intended to shoot Hank for his crime'' and Huo et al.'s system only extracts ``got'' from ``Hank got bitten by a snake''. Moreover, StoryFramer misses the actions ``summon'' and ``fall-in-love'' in the Aladdin story, which NaRuto finds.



\subsection{Expert Assessment of Action Models}
\label{sect:subjective}

As discussed above, it is difficult to align predicates between the ground truth and generated domain models; thus, one cannot say that generated action models have ``correct'' preconditions and effects by a simple comparison with ground truth. Instead, to evaluate the quality of our generated action models, we asked experts to rate the \emph{appropriateness} of each action's preconditions and effects, relative to the predicates that are defined in the respective domain model. We applied this evaluation to all four models of both domains, i.e., the ground truth domain model as well as the domain models generated by StoryFramer, H2020 and NaRuto, as described in section \ref{sect:baselines} above.

We recruited 9 participants, all of whom are experts in AI planning, and familiar with modelling planning domains in PDDL.
Each participant was given the four different models of one, or in a few cases both, of the domains, and asked to rate the appropriateness of each precondition and each effect of each action in all four domain versions, using a 5-point Likert scale: 1=not appropriate; 2=probably/maybe not appropriate; 3=undecided; 4=probably/maybe appropriate; 5=appropriate. The models were formatted to appear as similar as possible (e.g., comments were removed from the ground truth domain models, indentation was made uniform, etc). Participants were told only that all four domain models were ``automatically learned from narrative text'', and the order of presentation of the four models was randomized for each participant. We finally received $N$=$6$ responses for each of the two domains (stories).

From each response, we compute three summary measures: (1) the average ratings of all preconditions and effects within each domain model; (2) the percentage of ratings that are classified as in agreement (i.e., 4 or 5); and (3) the percentage of ratings that are in disagreement (i.e., 1 or 2), within each domain model. For each model variant, all measures are averaged over the $N$=$6$ responses for each domain.
The domain model generated by NaRuto in the evaluation is with full (global) negations, called NaRuto$_{(G)}$. We calculate measures for the version with restricted (local) negations, called NaRuto$_{(L)}$, by omitting the negated preconditions and effects that would not be present in this model.

\begin{table}
\begin{center}
\caption{The average scores over all the respondents for all actions' preconditions and effects within each domain model (\textbf{Sco.}); and the average percentage in agreement(\textbf{Agg.}) and disagreement (\textbf{Disagg.}) scores. \textbf{Type} indicates if the domain model is generated manually or by a semi-automated or (fully) automated system. \textbf{Bold numbers} indicates the best results and \underline{underline} denotes the second-best results.}
\scalebox{0.96}{
\begin{tabular}{llccc} 

\textbf{Method}& \textbf{Type}&\textbf{Sco. $\uparrow$} & \textbf{Agg. $\uparrow$} & \textbf{Disagg. $\downarrow$} \\
\toprule
\multicolumn{5}{l}{Domain 1: An Old American West Story}\\
\midrule

 GT~\cite{ware2014plan}  &Manual      & 4.34  & 82.7\% & 10.7\%\\
 \cline{2-5}
 
StoryFramer~\cite{hayton:etal:KEPS:2017} &Semi-auto       & \underline{3.26} &42.5\%  & \underline{26.3\%} \\
H2020~\cite{hayton:etal:AAAI:2020}   &Auto    & 2.54 & 17.7\% & 41.2\% \\

NaRuto$_{(G)}$ &Auto     &2.98  & \underline{43.0\%} & 39.3\% \\
NaRuto$_{(L)}$ &Auto    & \textbf{3.57} & \textbf{60.8\%} &  \textbf{25.3\%}\\

\midrule
\multicolumn{5}{l}{Domain 2: The Tale of Aladdin}\\
\midrule
  GT~\cite{riedl2010narrative} &Manual       &4.84  & 97.3\% & 1.2\%\\
 \cline{2-5}
StoryFramer~\cite{hayton:etal:KEPS:2017}  &Semi-auto     &\textbf{4.04}  & \textbf{74.8\%} & \textbf{17.3\%} \\
H2020~\cite{hayton:etal:AAAI:2020} &Auto      & 2.81 & 38.2\% & 48.8\% \\

NaRuto$_{(G)}$  &Auto   &3.03  & 41.0\% & 38.3\% \\
NaRuto$_{(L)}$  &Auto    & \underline{3.34} & \underline{52.0\%} & \underline{29.5\%}\\
[1.5pt]
\hline

\end{tabular}}

\label{tab-subjective-summary}
\end{center} 
\end{table}

\begin{figure}[!t]
\centering
\includegraphics[width=0.48\textwidth]{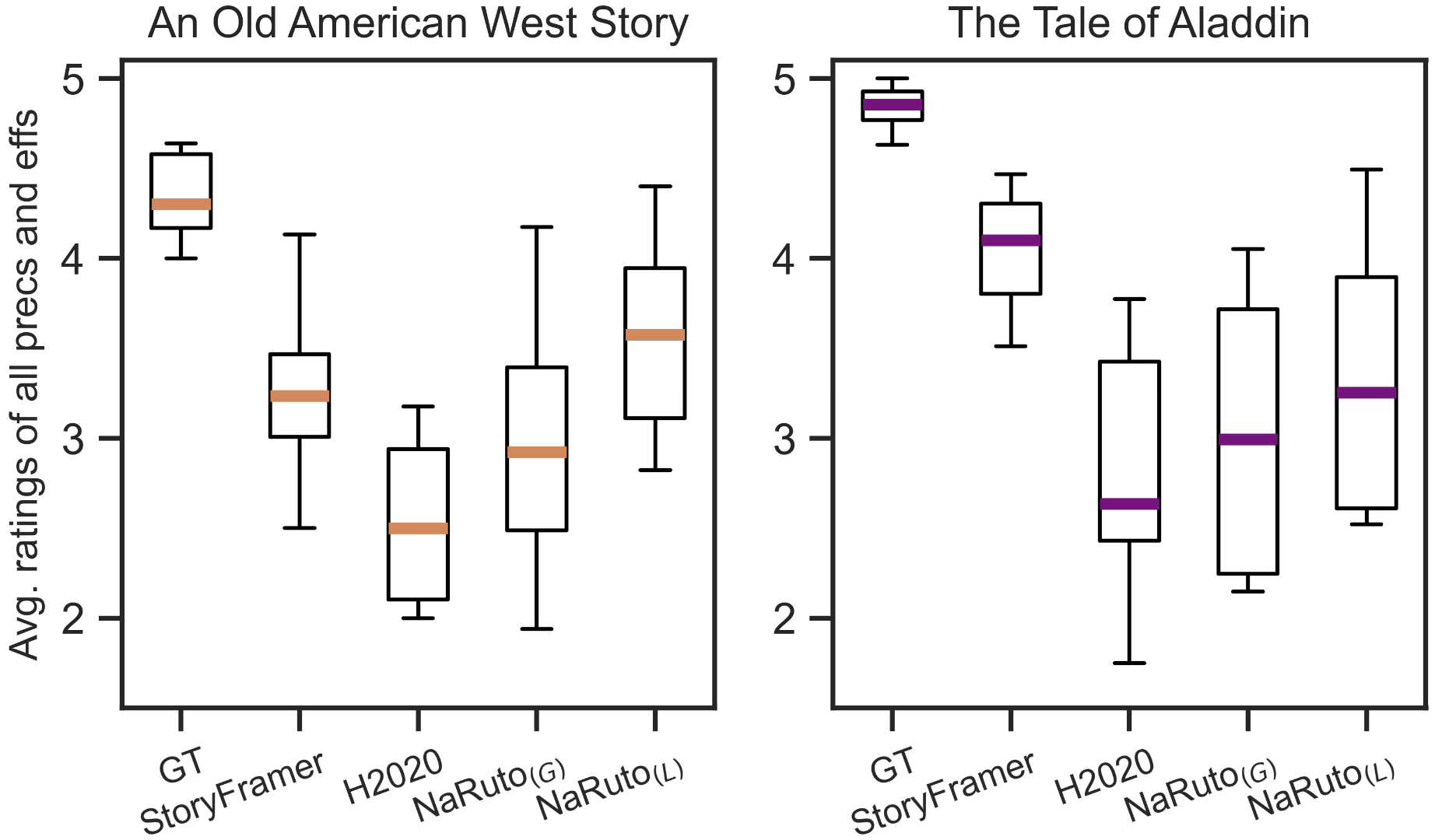}
\caption{Distribution, over respondents, of the average scores for all actions' preconditions and effects within each domain model. The thick line shows the median, box shows the interquartile range. Whiskers extend to the full range of values.}
\label{fig-subjective-avg-score}
\end{figure}

Results are summarized in Table \ref{tab-subjective-summary}.
Box-and-whiskers plots showing the distribution of average scores across responses for both of the domains are in Figure \ref{fig-subjective-avg-score}.

Unsurprisingly, the hand-written ground truth domain models receive the highest average and percentage-in-agreement scores, as well as the lowest percentage-in-disagreement scores.
The StoryFramer domain models also score well on both measures. Again, this is not surprising, since the selection of each action's preconditions and effects in these domain models was done manually (and presumably by a user with knowledge of the story they intend to model).
However, using NaRuto$_{(L)}$, our generated model of the Old American West domain is rated better than the StoryFramer model and our model of the Aladdin domain is rated second-best. This indicates that the precondition and effects predictor captures well the commonsense knowledge of actions in these domains.
The global negations strategy (NaRuto$_{(G)}$) is intended to capture the ramifications of positive action effects. However, the domain model with restricted negations (NaRuto$_{(L)}$) scores consistently better, indicating that most of the additional negated effects, and preconditions, in the global model are not helpful.
The domain models generated by H2020 score lower on both measures, indicating that the somewhat particular structure it encodes, which captures the sequence of events in the original story, is not perceived by experienced domain modellers as appropriate for a planning domain, which is normally expected to allow for the composition of all valid action sequences.

We also note that in all generated models, the average rating of actions' preconditions is consistently higher than that of their effects, sometimes significantly so (3.61\% to 74.80\%). This suggests generating appropriate effects is a harder problem.


\section{Conclusion and Future Work}

Narrative text exhibits many complexities, but it is also a rich source of knowledge about events and actions.
We presented the NaRuto system for creating planning-language-style action models from narrative texts. In contrast to previous approaches, our system does not depend on manual input, and creates action models that generalize beyond the event sequence in the source text.
Ultimately, this will enable the generation of planning models at a scale to support open-world, creative narrative planning.
In evaluation, our generated action models are rated better than those by comparable fully automated methods, and sometimes even than those by semi-automated systems using manual input or correction.
Yet, there are several areas for directing our future work: The commonsense relation predictor considers only two event arguments -- subject and object -- which currently limits action parameters to two. Our evaluation also revealed that selecting appropriate negative action effects, and taking story context into account when predicting precondition/effects, remain challenging tasks.

\bibliography{ecai}


\appendix

\section*{Appendix A\quad Evaluation for Argument Event Detection}

As described in section 3.1, the structured event representation construction part contains several individual components. Specifically, we evaluate the two pivotal and innovative components introduced in our approach: argument event detection and condition event detection.

Since both the Old American West story and the tale of Aladdin are short and relatively not as complex as other narrative texts, such as movie plots, only a few pairs of events from the sentences have a containing-contained relation. In order to test the general applicability of our designed argument detection rules (described in section 3.1.4), we performed a small-scale evaluation of this method. We manually annotated 114 pairs of events with containing-contained relations, which were extracted from 35 randomly selected sentences within a Movie plot summary dataset from Bamman et al.~\cite{bamman:movieplot:2013}, finding in 34 cases the contained event is indeed an argument of the containing event.
Based on this sample, the proposed rules achieve a precision of $1$ (i.e., no false positive) but a recall of $0.44$; thus, they are somewhat conservative but precise.

\section*{Appendix B\quad Evaluation for Condition Event Detection}

The problem of detecting condition--consequence relations between events
in texts has been studied, motivated in particular by finding causal
relationships \cite{puente:ICFS:2010}.
We evaluated two recent methods that detect conditional
structures, due to Fischbach et al.~\cite{fischbach:etal:2021} and Tan et al.~\cite{tan:CNC:2022},
respectively.
Both are BERT-based neural networks, but trained with different data.
Fischbach et al.\ use an annotated set of requirements documents
\cite{fischbach:dataset:2020}, while Tan et al.\ annotated and used a
set of news articles, together with the Penn Discourse Treebank 3.0
\cite{penn:journal:2019} and CausalTimeBank \cite{mirza:CTB:2014}
datasets. However, we also note that both are intended to extract
causal relations between events, which do not always coincide with
the condition--consequence relation.

We apply these two systems to the same set of 100 randomly selected
sentences from the Movie plot summary dataset.
Recall that these were selected to include the five signal words or phrases that we use (20 for each) and that 75 of them contain conditionals. 3 sentences have more than one condition--consequence event pair.
Both systems detect the presence of conditionals in a sentence in more cases than our method (59 and 60 of the 75 positive cases, respectively, compared to 53 for our method), but also have a much higher number of false positives (20 and 15 of the 25 negative cases, respectively, compared to 4 for our method),
leading to their lower precision, as shown in Table~\ref{tab-result-conditional}.
Furthermore, in true positive cases identified by each, we compare the events identified as conditions and consequences with our annotation.
These results are worse: Tan et al.'s system identifies the correct
text spans in only 6 of the 60 cases (EM-rate=0.1), while Fischbach et al.'s does so in 24 of the 59 cases.
On the other hand, our method is blind to any conditional expression
that does not use one of the five signal words or phrases.
(For example, the sentences ``Go away or I'll call the police!'' and
``Come back and I'll call the police!'' both express conditional
using conjunction, while ``I'll call the police and come back'' does not.)
We contend that further investigation into this particular aspect of event relationships is merited.

\begin{table}
\small
\begin{center}
\caption{Precision and recall of detecting the existence of conditionals in sentences. EM-rate is the proportion of sentences in which the detected condition and consequence events exactly match our annotation.}
\begin{tabular}{lccc} 

\textbf{} & \textbf{Precision} & \textbf{Recall} &\textbf{EM-rate} \\
\hline
Fischbach et al. (2021) & 0.75 & 0.79 & 0.41 \\

Tan et al. (2022)  & 0.80 & \textbf{0.80} & 0.1 \\
Ours. & \textbf{0.93} & 0.71 & \textbf{0.85} \\

\hline
\end{tabular}

\label{tab-result-conditional}
\end{center} 
\end{table}

\section*{Appendix C\quad Commonsense Knowledge Predictor Selection}
For the precondition and effect prediction in section 3.2.1, we investigated not only BART but also various other publicly available and freely accessible generative large language models: GPT2-Medium, GPT2-Large, GPT2-Extra Large \cite{radford2019language}, and T5 \cite{2020t5}), to fine-tune the COMET model. We assess their performance on the ATOMIC-2020 test set (a subset comprising exclusively the five selected relation tuples) by employing various well-established evaluation metrics pertinent to text generation: ROUGE-L \cite{lin2004rouge}, METEOR \cite{banerjee2005meteor} and BERT Score \cite{zhang2019bertscore}. For the hyper-parameter tuning of each model, we set the candidate sets of the learning rate, batch size and the number of epochs as [$1\mathrm{e}{-5}$, $5\mathrm{e}{-5}$], [16, 32, 64], and [2,3,4], respectively.
The results that achieve the best performance of each model are shown in Table~\ref{tab-comet-compare}. 
The BART model outperforms all the others, thus providing the rationale for our selection.

\begin{table}
\small
\begin{center}
\caption{The evaluation results of different metrics on 
 the generated commonsense knowledge inferences using different fine-tune models and the same beam search algorithm.}
\begin{tabular}{lccc} 
\textbf{} & \textbf{ROUGE-L} & \textbf{METEOR} & \textbf{BERT Score}\\
\hline
GPT2-M   & 0.41 & 0.30 & 0.57\\
GPT2-L   & 0.45 & 0.31 & 0.58\\
GPT2-XL  & 0.45 & 0.34 & 0.64\\
T5  & 0.48 & 0.36 & 0.64\\
BART (Ours.)  & \textbf{0.50} & \textbf{0.37} & \textbf{0.68}\\
\hline

\end{tabular}

\label{tab-comet-compare}
\end{center} 
\end{table}

\section*{Appendix D\quad Evaluation on Action Parameter Extraction}

In the action model creation phase of NaRuto, we proposed an event SRL argument name-based algorithm to extract the subject and object of the event as the parameters of the action (cf.\ section 3.2.3). To evaluate its efficacy, we randomly select 120 events from the Movie plot summary dataset. Among them, 65 events have no contained or argument event and 55 have argument events detected. For all 120 events, we manually annotate the parameters for evaluation. For the 65 events, the accuracy is 46/65=70.8\% and the number is 65.5\% for the 55 cases. Furthermore, as we observed, in most of the cases, the event's <subject, object> are as <ARG0, ARG1>, or <ARG1, ARG2>, or <ARG1, none>, or <ARG0, none> (for instance, over 78.3\% of all the events' parameters in the Movie plot summary dataset are determined as one of the 4 combinations by our algorithm). The precision on the 60 cases that are predicted as one of the 4 mentioned combinations is 41/60=68.3\%. 
These findings suggest that our algorithm possesses potential for further refinement, as it faces more challenges in detecting parameters within the more intricate main-event-contain-argument-event structure.


\end{document}